\documentclass{article}
\usepackage{spconf,amsmath,graphicx}
\usepackage{xcolor}

\newcommand{\Anurag}[1]{{\color{red}{\bf Anurag: }#1}}

\usepackage[
backend=biber,
style=ieee,
doi=false,isbn=false,url=false,eprint=false
]{biblatex}

\defbibheading{bibliography}[\refname]{}

\title{TAPLoss: A Temporal Acoustic Parameter Loss for Speech Enhancement}
\name{\begin{tabular}{c}Yunyang Zeng$^{1 \dagger}$, Joseph Konan$^{1 \dagger}$, Shuo Han$^{1 \dagger}$, David Bick$^{1 \dagger}$, Muqiao Yang$^{1 \dagger}$, \\Anurag Kumar$^2$, Shinji Watanabe$^1$, Bhiksha Raj$^1$ \thanks{$^\dagger$Equal Contribution (Random Order)} \end{tabular}}

\address{   
$^1$ Carnegie Mellon University, $^2$Meta Reality Labs Research }
\begin{document}
\ninept
\maketitle
\begin{abstract}

\noindent 
Speech enhancement models have greatly progressed in recent years, but still show limits in perceptual quality of their speech outputs. We propose an objective for perceptual quality based on temporal acoustic parameters. These are fundamental speech features that play an essential role in various applications, including speaker recognition and paralinguistic analysis. We provide a differentiable estimator for four categories of low-level acoustic descriptors involving: frequency-related parameters, energy or amplitude-related parameters, spectral balance parameters, and temporal features. 
Unlike prior work that looks at aggregated acoustic parameters or a few categories of acoustic parameters, our temporal acoustic parameter (TAP) loss enables auxiliary optimization and improvement of many fine-grain speech characteristics in enhancement workflows. We show that adding TAPLoss as an auxiliary objective in speech enhancement produces speech with improved perceptual quality and intelligibility. We use data from the Deep Noise Suppression 2020 Challenge to demonstrate that both time-domain models and time-frequency domain models can benefit from our method.

\end{abstract}
\section{Introduction}
\label{sec:intro}


Speech enhancement is aimed at enhancing the quality and intelligibility of degraded speech signals. The need for this arises in a variety of speech applications. While noise suppression or removal is an important part of the speech enhancement, retaining the perceptual quality of the speech signal is equally important. In recent years, deep neural networks based approaches have been the core of most state-of-the-art speech enhancement systems, in particular single channel speech enhancement \cite{10.1007/978-3-319-22482-4_11, https://doi.org/10.48550/arxiv.1703.09452, https://doi.org/10.48550/arxiv.1706.07162,
7364200}. These are traditionally trained using point-wise differences in time-domain or time-frequency-domain. 

However, many studies have shown limitations in these losses, including low correlations with speech quality \cite{perceptual_jnd} and \cite{metric_gan_plus}. Other studies have shown that they have overemphasis on high-energy phonemes \cite{plantinga2021perceptual} and an inability to improve pitch \cite{turian_henry}, resulting in speech that has artifacts or poor perceptual quality \cite{dnsmos_pesq_flaw}. The insufficiency of these losses has led to much work devoted to improving the perceptual quality of enhanced signals, which our work also aims to improve. Perceptual losses have often involved estimating the standard evaluation metrics such as Perceptual Evaluation of Speech Quality (PESQ) \cite{PESQ}. However, PESQ is non-differentiable, which forces difficult optimization \cite{fu2021metricgan} and often leads to limited improvements \cite{white_box_perceptual_loss, metric_RL_SE}. Other approaches use deep feature losses \cite{pase, pfpl_paper}; however, these have limited improvements because perceptual quality is only implicitly supervised. In this paper, we seek to address these issues by using fundamental speech features, which we refer to as acoustic parameters. 

The use of acoustic parameters has been shown to facilitate speaker classification, emotion recognition, and other supervised tasks involving speech characteristics \cite{sambur1975selection, brown1981experimental, tzirakis2017end}. Historically, these acoustic parameters were not incorporated in workflows with deep neural networks because they required non-differentiable computations. However, this does not reflect their significant correlation with voice quality in prior literature \cite{correlates_breathy_rough, correlates_breathy, KASUYA1986171}. Recently, some works have made progress in incorporating acoustic parameters for optimization of deep neural networks. Pitch, energy contour, and pitch contour were proposed to optimize perceptual quality in \cite{peng22d_interspeech}. However, these three parameters are a small subset of the characteristics we consider, and evaluation was not performed on standard English datasets. A wide range of acoustic parameters was proposed in \cite{yang22x_interspeech}, which introduced a differentiable estimator of these parameters to create an auxiliary loss aimed at forcing models to retain acoustic parameters. This proved to improve the perceptual quality of speech. Unlike prior work, which used summary statistics of acoustic parameters per utterance, our current estimator allows optimization \textit{at each time step}. The values of each parameter vary over time in the utterance, so statistics lose a significant amount of information in the comparison of clean and enhanced speech.

We look at 25 acoustic parameters -- \textbf{frequency related parameters}: pitch, jitter, F1, F2, F3 Frequency and bandwidth; \textbf{energy or amplitude-related parameters}: shimmer, loudness, harmonics-to-noise (HNR) ratio; \textbf{spectral balance parameters}: alpha ratio, Hammarberg index, spectral slope, F1, F2, F3 relative energy, harmonic difference; and additional \textbf{temporal parameters}: rate of loudness peaks, mean and standard deviation of length of voiced/unvoiced regions, and continuous voiced regions per second. We use OpenSmile \cite{10.1145/1873951.1874246} 
to perform the ground-truth non-differentiable calculations, creating a dataset to to train a differentiable estimator. 

Finally, we present our estimator for these 25 temporal acoustic parameters. Using the estimator, we define an acoustic parameter loss, coined TAPLoss, $\mathcal{L}_{\text{TAP}}$, that minimizes the distance between estimated acoustics for clean and enhanced speech. Unlike previous work, we do not assume the user has access to ground-truth clean acoustics. We empirically demonstrate the success of our method, observing improvement in relative and absolute enhancement metrics for perceptual quality and intelligibility.



\section{Methods}
\label{sec:method}
\begin{table*}[!htp]\centering
\scriptsize
\begin{tabular}{l|rrrrr|rrrrr|rrrrrr}\toprule
&\multicolumn{5}{c|}{Demucs $\mathcal{L}_{\text{TAP}}$ $\lambda_{1}$ Ablation ($\lambda_{2}$ = 0)} &\multicolumn{5}{c|}{Demucs $\mathcal{L}_{\text{TAP}}$ $\lambda_{2}$ Ablation ($\lambda_{1}$ = 1)} &\multicolumn{5}{c}{FullSubNet $\mathcal{L}_{\text{TAP}}$ $\gamma$ Ablation} \\ \hline
Weight &0.01 &0.03 &0.1 &0.3 &1 &0.01 &0.03 &0.1 &0.3 &1 &0.01 &0.03 &0.1 &0.3 &1 \\ \hline
PESQ &2.788 &2.841 &2.824 &2.834 &\textbf{2.859} &2.899 &2.903 &2.926 &\textbf{2.958} &2.958 &2.979 &\textbf{2.981} &2.979 &2.969 &2.965 \\
STOI &0.9697 &0.9698 &0.9689 &0.9689 &\textbf{0.9694} &0.9707 &0.9712 &0.9714 &\textbf{0.9722} &0.9720 &0.9654 &\textbf{0.9654} &0.9654 &0.9648 &0.9654 \\
\bottomrule
\end{tabular}
\caption{ $\mathcal{L}_{\text{TAP}}$ ablation study of Demucs hyperparameters, $\lambda_1$ and $\lambda_2$, and FullSubNet hyperparamter $\gamma$.}
\end{table*}

\begin{figure*}[!h]
    \begin{subfigure}{0.5\textwidth}
        \includegraphics[width=9cm]{
            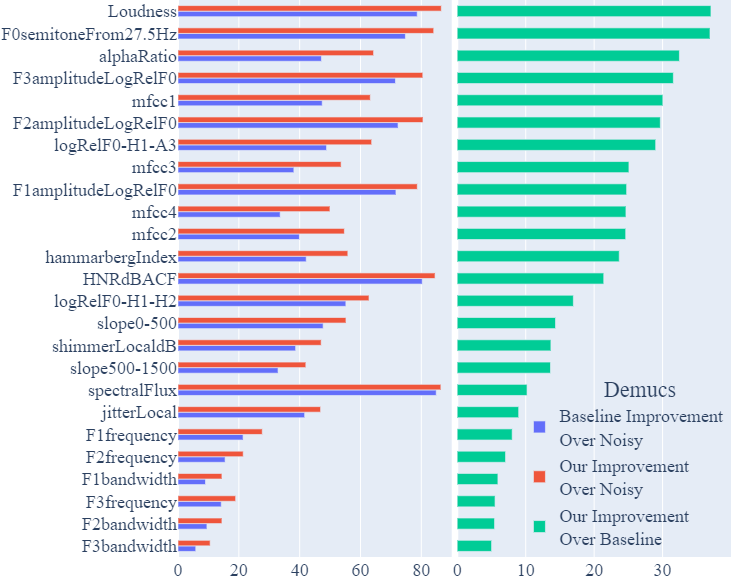}
    \end{subfigure}
    \begin{subfigure}{0.5\textwidth}
        \includegraphics[width=9cm]{
            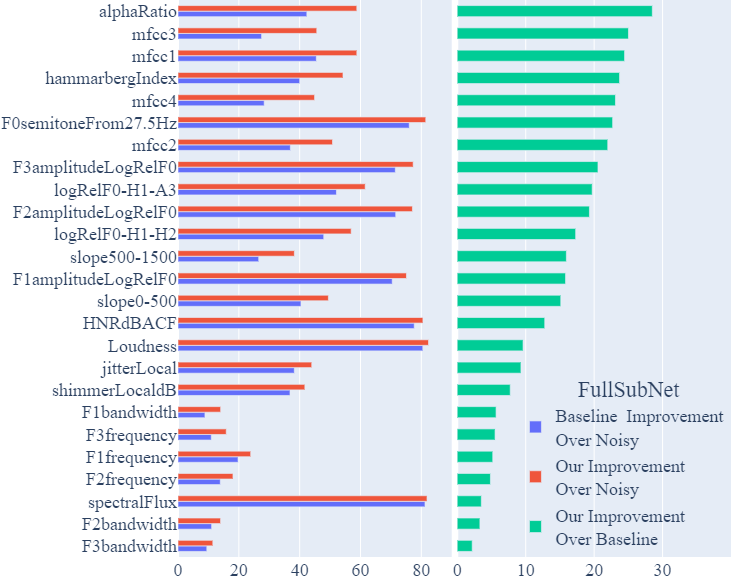}
    \end{subfigure}
\caption{Percent Acoustic Improvement $\mathbf{PAI}$ on DNS-2020 Synthetic Test (No Reverb). Compared are  
baseline improvement over noisy (blue) $\mathbf{PAI}(\mathbf{s_1}, \mathbf{x})$,
our improvement over noisy (red)
$\mathbf{PAI}(\mathbf{s_2}, \mathbf{x})$,
our improvement over baseline (green)
$\mathbf{PAI}(\mathbf{s_2}, \mathbf{s_1})$. 
Sorted by $\mathbf{PAI}(\mathbf{s_2}, \mathbf{s_1})$.}
\label{fig:relative_improvement}
\end{figure*}

\subsection{Background}
In the time domain, let $\mathbf{y}$ denote a signal with discrete duration $M$ such that $\mathbf{y} \in \mathbb{R}^M$.
We define clean speech signal $\mathbf{s}$, noise signal $\mathbf{n}$, and noisy speech signal $\mathbf{x}$ with the following additive relation:

\begin{equation}
\mathbf{x} = \mathbf{s} + \mathbf{n} \label{eq:se-td} \\
\end{equation}
Similarly, in the time-frequency domain, let $\mathbf{Y} \in \mathbb{C}^{T \times F}$ denote a complex spectrogram with $T$ discrete time frames and $F$ discrete frequency bins. $\mathfrak{Re}\{\mathbf{Y}\} \in \mathbb{R}^{T \times F}$ denotes real components and $\mathfrak{Im}\{\mathbf{Y}\} \in \mathbb{R}^{T \times F}$ denotes complex components. Let $Y(t,f)$ be the complex-valued time-frequency bin of $\mathbf{Y}$ at discrete time frame $t \in [0, T)$ and discrete frequency bin $f \in [0, F)$. By the linearity of Fourier transforms, the complex spectrograms for clean speech $\mathbf{S}$, noise $\mathbf{N}$, and noisy speech $\mathbf{X}$ relate with the additive relation:

\begin{equation}
    \mathbf{X} = \mathbf{S} + \mathbf{N} \label{eq:se-td} \\
\end{equation}
A speech enhancement model $G$ outputs enhanced signal $\mathbf{\hat{s}}$ such that:

\begin{equation}
    \left\{ 
        \mathbf{\hat{s}} = G(\mathbf{x}) \middle| \mathbf{\hat{s}}, \mathbf{x} \in \mathbb{R}^M
    \right\} 
\end{equation}
During optimization, $G$ minimizes the divergence between $\mathbf{s}$ and $\mathbf{\hat{s}}$. We denote $\mathbf{\hat{S}}$ to be enhanced complex spectrogram derived from $\mathbf{\hat{s}}$.

\subsection{Temporal Acoustic Parameter Estimator}





Let $\mathbf{A}_{\mathbf{y}} \in \mathbf{R}^{T \times 25}$ represent the 25 temporal acoustic parameters of signal $\mathbf{y}$ with $T$ discrete time frames. We represent $A_\mathbf{y}(t,p)$ as the acoustic parameter $p$ at discrete time frame $t$. We standardize the acoustic parameters to have mean 0 and variance 1 across the time dimension. Standardization helps optimization and analysis through consistent units across features. To predict $\mathbf{A}_{\mathbf{y}}$, we define estimator:

\begin{equation}
    \mathbf{\hat{A}}_{\mathbf{y}} = \mathcal{TAP}(\mathbf{y})
\end{equation}
$\mathcal{TAP}$ takes a signal input $\mathbf{y}$, derives complex spectrogram $\mathbf{Y}$ with $F=257$ frequency bins, and then passes the  complex spectrogram to a recurrent neural network to output the temporal acoustic parameter estimates $\mathbf{\hat{A}}_{\mathbf{y}}$.

For loss calculation, we define total mean absolute error as:

\begin{equation}
    \text{MAE} (\mathbf{A}_{\mathbf{y}}, \mathbf{\hat{A}}_{\mathbf{y}}) =
    \frac{1}{TP} \sum_{t=0}^{T-1} \sum_{p=0}^{P-1} | A_\mathbf{y}(t,p)-A_\mathbf{\hat{y}}(t,p) |
    \in \mathbb{R}
\end{equation}
During training, $\mathcal{TAP}$ parameters learn to minimize the divergence of 
$\text{MAE} ( \mathbf{A_s} , \mathbf{A_{\hat{s}}} )$
using Adam optimization.



\subsection{Temporal Acoustic Parameter Loss}

We developed temporal acoustic parameter loss, $\mathcal{L}_{\text{TAP}}$, to enable divergence minimization between clean and enhanced acoustic parameters. This section expounds the mathematical formulation of $\mathcal{L}_{\text{TAP}}$.



Let magnitude spectrogram $||\hat{S}(t,f)||$ represent the magnitude of complex spectrogram $\mathbf{\hat{S}}$. Using Parseval's Theorem, the frame energy weights, $\boldsymbol{\omega}$, is derived from the magnitude spectrogram mean across the frequency axis:

\begin{equation} \label{eq:energy-weight}
\boldsymbol{\omega} = \frac{1}{F} \sum_{f=0}^{F-1} ||\hat{S}(t,f)||^{2}\; \in \mathbb{R}^{T}
\end{equation}
Because high energies are perceived more noticeably, we apply sigmoid, $\sigma$, to emulate human hearing with bounded scales, resulting in smoothed energy weights $\sigma(\omega)$.

Finally, we define our temporal acoustic parameter loss, $\mathcal{L}_{\text{TAP}}$, as the mean absolute error between clean and enhanced acoustic parameter estimates with smoothed frame energy-weighting:

\begin{align}
&\mathcal{L}_{\text{TAP}} (\mathbf{s}, \mathbf{\hat{s}}) 
= \text{MAE} \left(
    \mathcal{TAP} (\mathbf{s})       \odot \sigma(\boldsymbol{\omega}), 
    \mathcal{TAP} (\mathbf{\hat{s}}) \odot \sigma(\boldsymbol{\omega})
    \right)
\end{align}
Here, "$\odot$" denotes elementwise multiplication with broadcasting.
Note that this loss is end-to-end differentiable and takes only waveform as input. Therefore, this loss enables acoustic optimization of \textit{any} speech model and task with clean references.

\section{Experiments}
\label{sec:exp}

    

\subsection{Workflow with TAPLoss}
This section describes the workflow with TAPLoss applied to speech enhancement models. To demonstrate that our method generalizes on both time-domain and time-frequency domain models, we apply the TAPLoss, $\mathcal{L}_{\text{TAP}}$ to two competitive SE models, Demucs \cite{defossez2020real} and FullSubNet \cite{hao2020fullsubnet}. Demucs is a mapping-based time domain model with an encoder-decoder structure that takes a noisy waveform as input and outputs an estimated clean waveform. FullSubNet is a masking-based time-frequency domain fusion model that combines a full-band and a sub-band model. FullSubNet estimates a complex Ideal Ratio Mask (cIRM) from the complex spectrogram of the input signal and multiplies the cIRM with the complex spectrogram of the input to get the complex spectrogram of the enhanced signal. The enhanced complex spectrogram translates to the time-domain through inverse short-time Fourier transform (i-STFT). 

Our goal is to fine-tune the two baseline enhancement models with $\mathcal{L}_{\text{TAP}}$ to improve their perceptual quality and intelligibility. During forward propagation, the enhancement model takes a noisy signal as input and outputs an enhanced signal. The TAP estimator predicts temporal acoustic parameters for both clean and enhanced signals. $\mathcal{L}_{\text{TAP}}$ is then computed through the methods discussed in the previous subsection. Demucs and FullSubNet also have their own loss functions. FullSubNet uses mean squared error (MSE) between the estimated cIRM and the true cIRM as loss ($\mathcal{L}_{\text{cIRM}}$). Demucs has two loss functions, L1 waveform loss ($\mathcal{L}_{\text{wave}}$) and multi-resolution STFT loss ($\mathcal{L}_{\text{STFT}}$). The baseline Demucs model pre-trained on the DNS 2020 dataset only uses L1 waveform loss. In order for a fair comparison, we first fine-tune Demucs using L1 waveform loss and $\mathcal{L}_{\text{TAP}}$. However, previous works have shown that Demucs model is prone to generating tonal artifacts \cite{https://doi.org/10.48550/arxiv.2111.11773} and we have observed this phenomenon during fine-tuning with L1 waveform loss and $\mathcal{L}_{\text{TAP}}$. Moreover, we discovered that the multi-resolution STFT loss could alleviate this issue because the error introduced by tonal artifacts is more significant and obvious in the time-frequency domain than in the time domain. Therefore, from the best fine-tuning result, we fine-tune again with L1 waveform loss, $\mathcal{L}_{\text{TAP}}$, and multi-resolution STFT loss to remove the tonal artifacts. The following equations show final loss functions for fine-tuning Demucs and FullSubNet, where $\lambda_{1}$, $\lambda_{2}$ and ${\gamma}$ denote weight hyperparameters:

\begin{align}
&\mathcal{L}_{\text{Demucs}} = \mathcal{L}_{\text{wave}} + \lambda_{1} \cdot \mathcal{L}_{\text{TAP}} + \lambda_{2} \cdot \mathcal{L}_{\text{STFT}} \\
&\mathcal{L}_{\text{FullSubNet}} = \mathcal{L}_{\text{cIRM}} + \gamma \cdot \mathcal{L}_{\text{TAP}}
\end{align}
During backward propagation, TAP estimator parameters are frozen and only enhancement model parameters are optimized.

\subsection{Data}
This study uses 2020 Deep Noise Suppression Challenge (DNS) data \cite{reddy2020interspeech}, which includes clean speech (from Librivox corpus), noise (from Freesound and AudioSet \cite{gemmeke2017audio}), and noisy speech synthesis methods. We synthesize thirty-second clean-noisy pairs, including 50,000 samples for training and 10,000 samples for development. In our experiments, we use the official synthetic test set with no reverberation, which has 150 ten-second samples.


\begin{table*}[!htp]\centering
\scriptsize
\begin{tabular}{llrrrrrrrrrrrr}\toprule
Metric &Loss(es) used &NB PESQ &WB PESQ &STOI &ESTOI &CD &LLR &WSS &OVRL &BAK &SIG &NORESQA \\ \toprule
Clean &-- &-- &-- &-- &-- &-- &-- &-- &\textbf{3.28} &\textbf{4.04} &\textbf{3.56} &\textbf{4.61} \\ 
Noisy &-- &2.454 &1.582 &0.915 &0.810 &12.623 &0.577 &35.546 &2.48 &2.62 &3.39 &2.99 \\ \toprule
Demucs &$\mathcal{L}_{\text{wave}}$ &3.272 &2.652 &0.965 &0.921 &17.138 &0.443 &18.239 &3.31 &\textbf{4.15} &3.54 &3.95\\
Demucs &$\mathcal{L}_{\text{wave}} + \lambda_{1} \mathcal{L}_{\text{TAP}}$ &3.356 &2.859 &0.969 &0.930 &17.803 &0.334 &23.442 &3.15 &3.78 &\textbf{3.58} &\textbf{4.12} \\
Demucs &$\mathcal{L}_{\text{wave}} + \lambda_{1} \mathcal{L}_{\text{TAP}} + \lambda_{2} \mathcal{L}_{\text{STFT}}$ &\textbf{3.409} &\textbf{2.958} &\textbf{0.972} &\textbf{0.934} &\textbf{18.298} &\textbf{0.312} &\textbf{14.392} &\textbf{3.34} &4.14 &3.57 &4.08 \\ \toprule
FullSubNet &$\mathcal{L}_{\text{cIRM}}$ &3.386 &2.889 &0.964 &0.920 &16.962 &0.399 &20.887 &3.21 &4.02 &3.51 &4.09 \\
FullSubNet &$\mathcal{L}_{\text{cIRM}} + \gamma \mathcal{L}_{\text{TAP}}$ &\textbf{3.417} &\textbf{2.981} &\textbf{0.965} &\textbf{0.922} &\textbf{17.677} &\textbf{0.310} &\textbf{18.946} &\textbf{3.25} &\textbf{4.05} &\textbf{3.53} &\textbf{4.14} \\
\bottomrule
\end{tabular}
\caption{Relative and absolute measures of speech enhancement quality, comparing $\mathcal{L}_{\text{TAP}}$ with the baseline on DNS-2020 Test (No Reverb).}\label{tab: metrics}
\end{table*}

\begin{figure*}[!h]
    \begin{subfigure}{0.24\textwidth}
        \includegraphics[width=4.23cm]{
            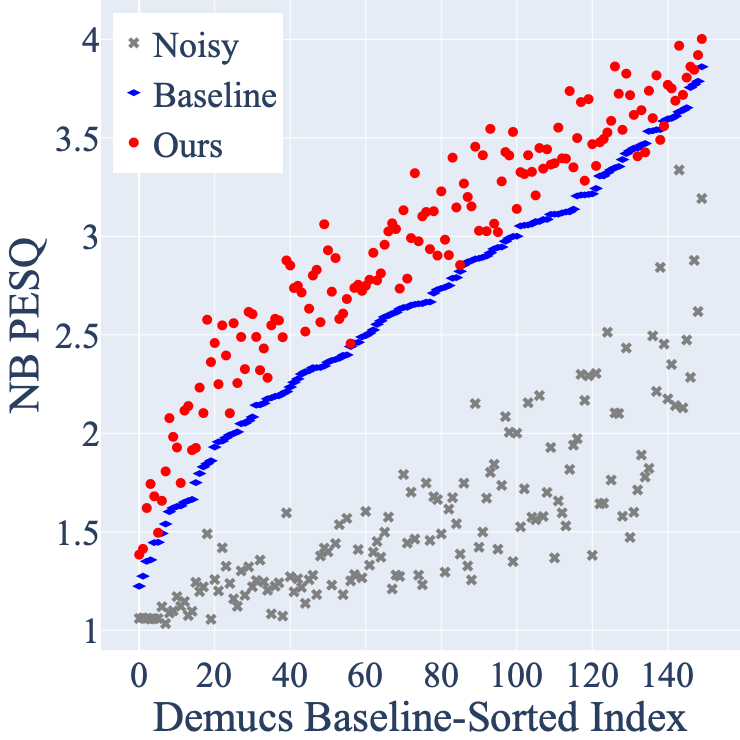}
    \end{subfigure}
    \begin{subfigure}{0.24\textwidth}
        \includegraphics[width=4.23cm]{
            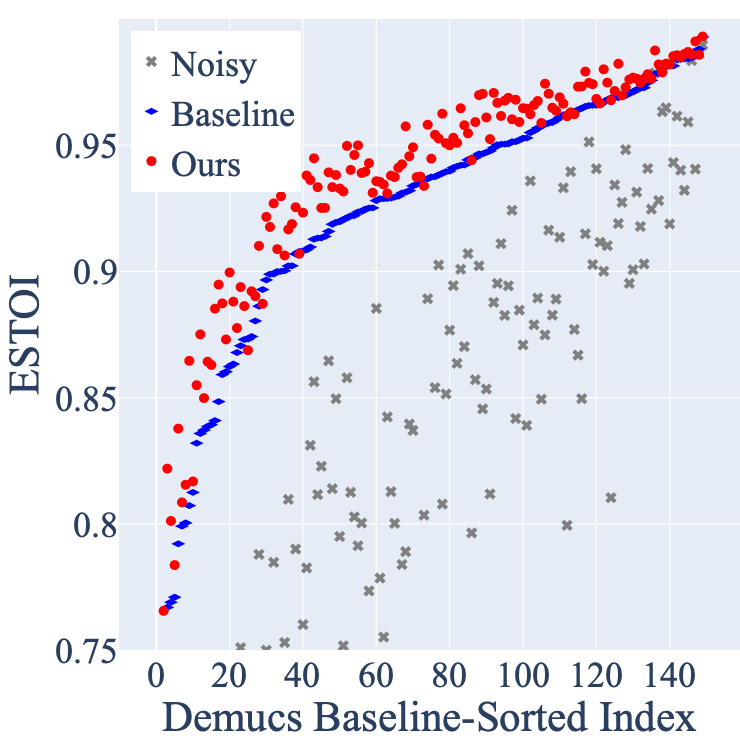}
    \end{subfigure}
    \begin{subfigure}{0.24\textwidth}
        \includegraphics[width=4.23cm]{
            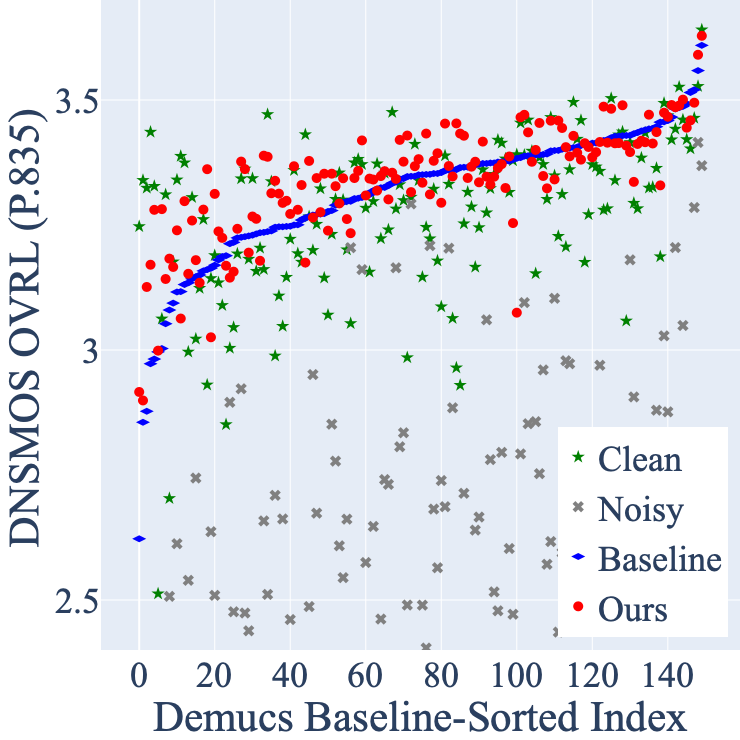}
    \end{subfigure}
    \begin{subfigure}{0.24\textwidth}
        \includegraphics[width=4.23cm]{
            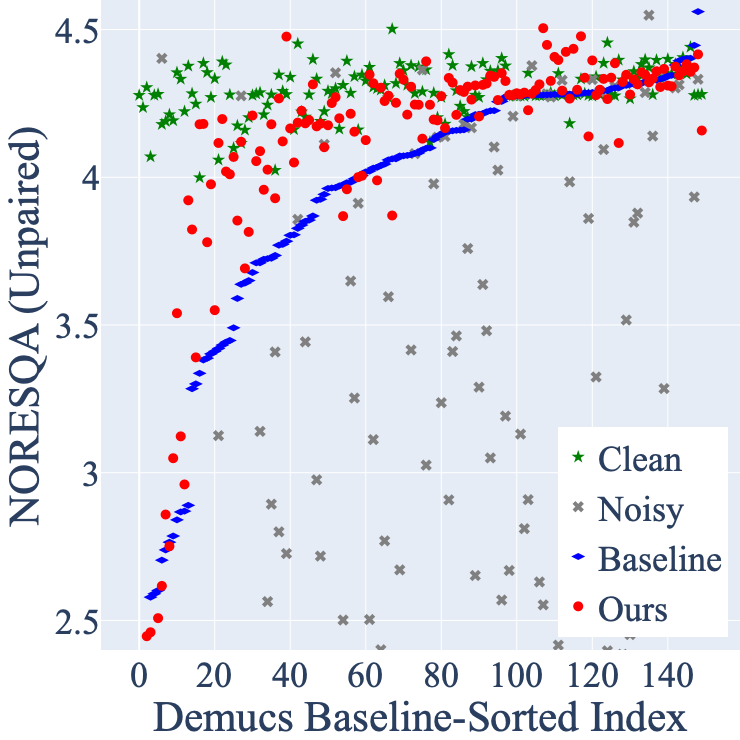}
    \end{subfigure}
\caption{Pairwise comparison of selected relative and absolute metrics with final $\mathcal{L}_{\text{TAP}}$ Demucs with baseline on DNS-2020 Test (No Reverb).}
\label{fig:comparison}
\end{figure*}


\subsection{Experiment Details And Ablation}
We fine-tune official pre-trained checkpoints of Demucs \footnote{\url{https://github.com/facebookresearch/denoiser}} and FullSubNet \footnote{\url{https://github.com/haoxiangsnr/FullSubNet}}. 
We fine-tune Demucs for 40 epochs with acoustic weight $\lambda_{1}$ and another 10 epochs with spectrogram weight $\lambda_2$. We fine-tune FullSubNet for 100 epochs with acoustic weight $\gamma$. TAP and TAPLoss source code will be available at \url{https://github.com/YunyangZeng/TAPLoss}. In our experiments, the estimator architecture is a 3-layer bidirectional recurrent neural network with long short-term memory (LSTM), with 256 hidden units. After 200 epochs, the acoustic parameter estimator converges to a training error of 0.15 and a validation error of 0.15.

As an auxiliary loss, $\mathcal{L}_{\text{TAP}}$ requires ablation to determine an optimal hyperparameter specification. We observe that Demucs benefits from high acoustic weights in the time-domain model. However, we also observed tonal artifacts when listening to the audio and visualizing the spectrogram. Upon investigation, these artifacts were caused by the model's architecture and are a known issue with some transpose convolution specifications. To address tone issues, we performed another ablation to improve spectrograms, given the acoustic weight. We observed that high spectrogram weights help, but it is not as important as optimizing the acoustics. In the time-frequency domain, 0.03 as an acoustic weight gave the best result on FullSubNet. Notably, a weight of 1 for Demucs and a weight of 0.03 for FullSubNet account for respective scale differences.

\subsection{Acoustic Evaluation}



Consider a clean speech target $\mathbf{s}$, noisy speech input $\mathbf{x}$, baseline enhanced speech $\mathbf{\hat{s}_1}$, and our enhanced speech $\mathbf{\hat{s}_2}$. Let $\mathbf{MAE}_0$ denote the mean absolute error across time (axis 0), and let $\oslash$ represent element-wise division. We define percent acoustic improvement, $\mathbf{PAI}$, as follows:

\begin{align}
    \mathbf{\epsilon}_{\mathbf{m}, \mathbf{m}^\prime}
        & \triangleq 
            \mathbf{MAE}_0(\mathbf{A}_\mathbf{m}, 
                \mathbf{A}_{\mathbf{m}^\prime}) \\
    \mathbf{PAI}(\mathbf{A}_\mathbf{\hat{s}_1}, \mathbf{A}_\mathbf{x}) 
        & = 100\% \cdot \left( 1 - 
        \mathbf{\epsilon}_{\mathbf{\hat{s}_1},\mathbf{s}} \oslash
        \mathbf{\epsilon}_{\mathbf{x},\mathbf{s}} \right) \\
    \mathbf{PAI}(\mathbf{A}_\mathbf{\hat{s}_2}, \mathbf{A}_\mathbf{x})
        & = 100\% \cdot \left( 1 - 
        \mathbf{\epsilon}_{\mathbf{\hat{s}_2},\mathbf{s}} \oslash
        \mathbf{\epsilon}_{\mathbf{x},\mathbf{s}} \right) \\
    \mathbf{PAI}(\mathbf{A}_\mathbf{\hat{s}_2}, \mathbf{A}_\mathbf{\hat{s}_1})
        & = 100\% \cdot \left( 1 - 
        \mathbf{\epsilon}_{\mathbf{\hat{s}_2},\mathbf{s}} \oslash
        \mathbf{\epsilon}_{\mathbf{\hat{s}_1},\mathbf{s}}\right) 
\end{align}
Acoustic evaluation involves three components: (1) baseline improvement over noisy speech input, $\mathbf{PAI}(\mathbf{A}_\mathbf{\hat{s}_1}, \mathbf{A}_\mathbf{x})$, (2) our improvement over noisy speech input, $\mathbf{PAI}(\mathbf{A}_\mathbf{\hat{s}_2}, \mathbf{A}_\mathbf{x})$, and (3) our improvement compared to the baseline, $\mathbf{PAI}(\mathbf{A}_\mathbf{\hat{s}_2}, \mathbf{A}_\mathbf{\hat{s}_1})$. 

Acoustic improvement measures how well enhancement processes noisy inputs into clean-sounding output. 0\% improvement means enhancement has not changed noisy acoustics, while 100\% is maximum possible improvement with enhanced acoustics identical to clean acoustics. 
Relative acoustic improvement measures how well enhancement fine-tuning yields a more clean-sounding output. 0\% improvement means TAPLoss has not changed enhanced acoustics after fine-tuning. 100\% improvement means TAPLoss enhanced acoustics sound identical to clean acoustics.

Figure \ref{fig:relative_improvement} presents percentage acoustic improvement for Demucs and FullSubNet. On average, Demucs with $\mathcal{L}_{\text{wave}}$, $\mathcal{L}_{\text{TAP}}$ and $\mathcal{L}_{\text{STFT}}$ improved noisy acoustics by 53.9\% while the baseline Demucs improved them by 44.9\%. FullSubNet with $\mathcal{L}_{\text{cIRM}}$, $\mathcal{L}_{\text{TAP}}$ improved noisy acoustics by 50.3\% while the baseline FullSubNet improved them by 42.6\%. 
 On average, TAPLoss improved Demucs baseline acoustics by 19.4\% and FullSubNet baseline acoustics by 14.5\%.

As an analytic tool, acoustics decompose enhancement quality changes -- identifying potential architectural or optimization criteria in need of development. For example, Demucs and FullSubNet architectures demonstrate difficulty optimizing formant frequency and bandwidth. As such, these empirical results suggest that future work introducing related digital signal processing mechanisms could enable improved acoustic fidelity optimization capacity. By providing a framework for acoustic analysis and optimization, this paper provides the tools needed to understand and improve acoustics and perceptual quality.

\subsection{Perceptual Evaluation}

Enhancement evaluation includes relative metrics that compare signals, and absolute metrics that valuate individual signals. Relative metrics include Short-Time Objective Intelligibility (STOI), extended Short-Time Objective Intelligibility (ESTOI), Cepstral Distance (CD), Log-Likelihood Ratio (LLR), Weighted Spectral Slope (WSS), Wide-band (WB) and Narrow-band (NB) Perceptual Evaluation of Speech Quality (PESQ) \cite{10.5555/2484638}. Absolute metrics include overall (OVRL), signal (SIG), and background (BAK) from DNSMOS P.835 \cite{9746108}. Finally, Non-matching Reference based Speech Quality Assessment (NORESQA) includes its absolute (unpaired) and relative (paired) MOS estimates \cite{noresqa}.

Many enhancement evaluation metrics benefit from the explicit optimization of acoustic parameters using TAPLoss. Perceptual evaluation of speech quality, both narrow band (PESQ-NB) and wideband (PESQ-WB), improved most significantly. While STOI did not improve much in the time-frequency domain model, it improved modestly in the time-domain model. Improvement occurs in most DNSMOS metrics (OVRL, SIG, BAK) in time-frequency and time domain models; however, enhancement outperforming clean suggests the metric is unreliable. NORESQA saw significant gains in this time-domain application, though explicit spectrogram optimization hurts the metric given a source time-domain model. Based on this empirical analysis, we recommend TAPLoss in situations with perceptual quality improvement objectives. Future works may significantly benefit perceptual quality by weighing acoustics given a specific metric optimization objective. 

Figure \ref{fig:comparison} presents more details that help us analyze the 150 ten-second samples. In order to facilitate pairwise comparison, we rank order by the baseline enhanced speech evaluation. By comparison, our enhanced speech outperforms the baseline enhanced speech on the two relative metrics, NB-PESQ and ESTOI. A similar pattern can be observed while analyzing NORESQA. The NORESQA of our enhanced speech mostly outperforms baseline-enhanced speech.

\vspace{-2mm}
\section{Conclusion}

\vspace{-2mm}

TAPLoss can improve acoustic fidelity in both time domain and time-frequency domain speech enhancement models. In contrast to aggregated acoustic parameters, optimization of temporal acoustic parameters yield better enhancement evaluation and significantly better acoustic improvement. Further, acoustic improvement using TAPLoss has strong foundations in digital signal processing, informing tailored future developments of acoustically motivated architectural changes or loss optimizations to improve speech enhancement.

\vspace{-2mm}
\section{ACKNOWLEDGEMENT}
\vspace{-2mm}
This work used the Extreme Science and Engineering Discovery Environment (XSEDE) ~\cite{xsede}, which is supported by National Science Foundation grant number ACI-1548562. Specifically, it used the Bridges system ~\cite{nystrom2015bridges}, which is supported by NSF award number ACI-1445606, at the Pittsburgh Supercomputing Center (PSC).

\vfill\pagebreak


\section{References}
\printbibliography

@InProceedings{10.1007/978-3-319-22482-4_11,
author="Weninger, Felix
and Erdogan, Hakan
and Watanabe, Shinji
and Vincent, Emmanuel
and Le Roux, Jonathan
and Hershey, John R.
and Schuller, Bj{\"o}rn",
editor="Vincent, Emmanuel
and Yeredor, Arie
and Koldovsk{\'y}, Zbyn{\v{e}}k
and Tichavsk{\'y}, Petr",
title="Speech Enhancement with {LSTM} Recurrent Neural Networks and its Application to Noise-Robust {ASR}",
booktitle="Latent Variable Analysis and Signal Separation",
year="2015",
publisher="Springer International Publishing",
address="Cham",
pages="91--99",
isbn="978-3-319-22482-4"
}

@article{https://doi.org/10.48550/arxiv.1703.09452,
  title={{SEGAN}: Speech Enhancement Generative Adversarial Network},
  author={Pascual, Santiago and Bonafonte, Antonio and Serr{\`a}, Joan},
  journal={arXiv preprint arXiv:1703.09452},
  year={2017}
}

@inproceedings{https://doi.org/10.48550/arxiv.1706.07162,
  title={A wavenet for speech denoising},
  author={Rethage, Dario and Pons, Jordi and Serra, Xavier},
  booktitle={Proc. ICASSP},
  pages={5069--5073},
  year={2018},
  organization={IEEE}
}

@ARTICLE{7364200,
  author={Williamson, Donald S. and Wang, Yuxuan and Wang, DeLiang},
  journal={IEEE/ACM Transactions on Audio, Speech, and Language Processing}, 
  title={Complex Ratio Masking for Monaural Speech Separation}, 
  year={2016},
  volume={24},
  number={3},
  pages={483-492},
  doi={10.1109/TASLP.2015.2512042}}

@inproceedings{yang22x_interspeech,
  author={Muqiao Yang and Joseph Konan and David Bick and Anurag Kumar and Shinji Watanabe and Bhiksha Raj},
  title={{Improving Speech Enhancement through Fine-Grained Speech Characteristics}},
  year=2022,
  booktitle={Proc. Interspeech},
  pages={2953--2957},
  doi={10.21437/Interspeech.2022-11161}
}

@article{reddy2020interspeech,
  title={The {Interspeech} 2020 Deep Noise Suppression Challenge: Datasets, Subjective Testing Framework, and Challenge Results},
  author={Reddy, Chandan KA and Gopal, Vishak and Cutler, Ross and Beyrami, Ebrahim and Cheng, Roger and Dubey, Harishchandra and Matusevych, Sergiy and Aichner, Robert and Aazami, Ashkan and Braun, Sebastian and others},
  journal={Proc. Interspeech},
  year={2020}
}

@inproceedings{defossez2020real,
  title={Real Time Speech Enhancement in the Waveform Domain},
  author={Defossez, Alexandre and Synnaeve, Gabriel and Adi, Yossi},
  booktitle={Proc. Interspeech},
  year={2020}
}

@INPROCEEDINGS{hao2020fullsubnet,
    author={Hao, Xiang and Su, Xiangdong and Horaud, Radu and Li, Xiaofei},
    booktitle={Proc. ICASSP},
    title={Fullsubnet: A Full-Band and Sub-Band Fusion Model for Real-Time Single-Channel Speech Enhancement},
    year={2021},
    pages={6633-6637},
    doi={10.1109/ICASSP39728.2021.9414177}
}

@INPROCEEDINGS{9746108,
  author={Reddy, Chandan K A and Gopal, Vishak and Cutler, Ross},
  booktitle={Proc. ICASSP}, 
  title={{DNSMOS P}.835: A Non-Intrusive Perceptual Objective Speech Quality Metric to Evaluate Noise Suppressors}, 
  year={2022},
  volume={},
  number={},
  pages={886-890},
  doi={10.1109/ICASSP43922.2022.9746108}}

@inproceedings{noresqa,
  title={{NORESQA}: A Framework for Speech Quality Assessment using Non-Matching References},
  author={Pranay Manocha and Buye Xu and Anurag Kumar},
  booktitle={Thirty-Fifth Conference on Neural Information Processing Systems},
  year={2021},
  url={https://proceedings.neurips.cc/paper/2021/file/bc6d753857fe3dd4275dff707dedf329-Paper.pdf}
}

@article{https://doi.org/10.48550/arxiv.2111.11773,
  title={Upsampling layers for music source separation},
  author={Pons, Jordi and Serr{\`a}, Joan and Pascual, Santiago and Cengarle, Giulio and Arteaga, Daniel and Scaini, Davide},
  journal={arXiv preprint arXiv:2111.11773},
  year={2021}
}

@book{10.5555/2484638,
author = {Loizou, Philipos C.},
title = {Speech Enhancement: Theory and Practice},
year = {2013},
isbn = {1466504218},
publisher = {CRC Press, Inc.},
address = {USA},
edition = {2nd}
}

@ARTICLE{xsede,
author = {J. Towns and T. Cockerill and M. Dahan and I. Foster and K. Gaither and A. Grimshaw and V. Hazlewood and S. Lathrop and D. Lifka and G. D. Peterson and R. Roskies and J. R. Scott and N. Wilkins-Diehr},
journal = {Computing in Science \& Engineering},
title = {XSEDE: Accelerating Scientific Discovery},
year = {2014},
volume = {16},
number = {5},
pages = {62-74},
doi = {10.1109/MCSE.2014.80},
url = {doi.ieeecomputersociety.org/10.1109/MCSE.2014.80},
ISSN = {1521-9615},
month={Sept.-Oct.}
}

@inproceedings{nystrom2015bridges,
  title={Bridges: a uniquely flexible HPC resource for new communities and data analytics},
  author={Nystrom, Nicholas A and Levine, Michael J and Roskies, Ralph Z and Scott, J Ray},
  booktitle={Proceedings of the 2015 XSEDE Conference: Scientific Advancements Enabled by Enhanced Cyberinfrastructure},
  pages={1--8},
  year={2015}
}

@article{sambur1975selection,
  title={Selection of acoustic features for speaker identification},
  author={Sambur, Marvin},
  journal={IEEE Transactions on Acoustics, Speech, and Signal Processing},
  volume={23},
  number={2},
  pages={176--182},
  year={1975},
  publisher={IEEE}
}

@article{brown1981experimental,
  title={An experimental study of the relative importance of acoustic parameters for auditory speaker recognition},
  author={Brown, Roger},
  journal={Language and Speech},
  volume={24},
  number={4},
  pages={295--310},
  year={1981},
  publisher={Sage Publications Sage CA: Thousand Oaks, CA}
}

@article{tzirakis2017end,
  title={End-to-end multimodal emotion recognition using deep neural networks},
  author={Tzirakis, Panagiotis and Trigeorgis, George and Nicolaou, Mihalis A and Schuller, Bj{\"o}rn W and Zafeiriou, Stefanos},
  journal={IEEE Journal of selected topics in signal processing},
  volume={11},
  number={8},
  pages={1301--1309},
  year={2017},
  publisher={IEEE}
}

@inproceedings{10.1145/1873951.1874246,
author = {Eyben, Florian and W\"{o}llmer, Martin and Schuller, Bj\"{o}rn},
title = {Opensmile: The Munich Versatile and Fast Open-Source Audio Feature Extractor},
year = {2010},
isbn = {9781605589336},
publisher = {Association for Computing Machinery},
address = {New York, NY, USA},
url = {https://doi.org/10.1145/1873951.1874246},
doi = {10.1145/1873951.1874246},
booktitle = {Proceedings of the 18th ACM International Conference on Multimedia},
pages = {1459–1462},
numpages = {4},
location = {Firenze, Italy},
series = {MM '10}
}

@inproceedings{peng22d_interspeech,
  author={Chiang-Jen Peng and Yun-Ju Chan and Yih-Liang Shen and Cheng Yu and Yu Tsao and Tai-Shih Chi},
  title={{Perceptual Characteristics Based Multi-objective Model for Speech Enhancement}},
  year=2022,
  booktitle={Proc. Interspeech},
  pages={211--215},
  doi={10.21437/Interspeech.2022-11197}
}

@INPROCEEDINGS{PESQ,
  author={Rix, A.W. and Beerends, J.G. and Hollier, M.P. and Hekstra, A.P.},
  booktitle={Proc. ICASSP}, 
  title={Perceptual evaluation of speech quality ({PESQ})-a new method for speech quality assessment of telephone networks and codecs}, 
  year={2001},
  volume={2},
  number={},
  pages={749-752 vol.2},
  doi={10.1109/ICASSP.2001.941023}}

@inproceedings{pase,
  title={Perceptual loss based speech denoising with an ensemble of audio pattern recognition and self-supervised models},
  author={Kataria, Saurabh and Villalba, Jes{\'u}s and Dehak, Najim},
  booktitle={Proc. ICASSP},
  pages={7118--7122},
  year={2021},
  organization={IEEE}
}

@inproceedings{pfpl_paper,
  author={Tsun-An Hsieh and Cheng Yu and Szu-Wei Fu and Xugang Lu and Yu Tsao},
  title={{Improving Perceptual Quality by Phone-Fortified Perceptual Loss Using Wasserstein Distance for Speech Enhancement}},
  year=2021,
  booktitle={Proc. Interspeech},
  pages={196--200},
  doi={10.21437/Interspeech.2021-582}
}

@article{fu2021metricgan,
  title={Metricgan+: An improved version of metricgan for speech enhancement},
  author={Fu, Szu-Wei and Yu, Cheng and Hsieh, Tsun-An and Plantinga, Peter and Ravanelli, Mirco and Lu, Xugang and Tsao, Yu},
  journal={Proc. Interspeech},
  year={2021}
}

@ARTICLE{white_box_perceptual_loss,
  author={Martin-Doñas, Juan Manuel and Gomez, Angel Manuel and Gonzalez, Jose A. and Peinado, Antonio M.},
  journal={IEEE Signal Processing Letters}, 
  title={A Deep Learning Loss Function Based on the Perceptual Evaluation of the Speech Quality}, 
  year={2018},
  volume={25},
  number={11},
  pages={1680-1684},
  doi={10.1109/LSP.2018.2871419}}

@INPROCEEDINGS{metric_RL_SE,
  author={Koizumi, Yuma and Niwa, Kenta and Hioka, Yusuke and Kobayashi, Kazunori and Haneda, Yoichi},
  booktitle={Proc. ICASSP}, 
  title={DNN-based source enhancement self-optimized by reinforcement learning using sound quality measurements}, 
  year={2017},
  volume={},
  number={},
  pages={81-85},
  doi={10.1109/ICASSP.2017.7952122}}

@inproceedings{gemmeke2017audio,
  title={Audio set: An ontology and human-labeled dataset for audio events},
  author={Gemmeke, Jort F and Ellis, Daniel PW and Freedman, Dylan and Jansen, Aren and Lawrence, Wade and Moore, R Channing and Plakal, Manoj and Ritter, Marvin},
  booktitle={Proc. ICASSP},
  pages={776--780},
  year={2017},
  organization={IEEE}
}

@article{perceptual_jnd,
  title={A differentiable perceptual audio metric learned from just noticeable differences},
  author={Manocha, Pranay and Finkelstein, Adam and Zhang, Richard and Bryan, Nicholas J and Mysore, Gautham J and Jin, Zeyu},
  journal={Proc. Interspeech},
  year={2020}
}

@article{metric_gan_plus,
  title={Metricgan+: An improved version of metricgan for speech enhancement},
  author={Fu, Szu-Wei and Yu, Cheng and Hsieh, Tsun-An and Plantinga, Peter and Ravanelli, Mirco and Lu, Xugang and Tsao, Yu},
  journal={Proc. Interspeech},
  year={2021}
}

@article{turian_henry,
  title={I'm sorry for your loss: Spectrally-based audio distances are bad at pitch},
  author={Turian, Joseph and Henry, Max},
  journal={arXiv preprint arXiv:2012.04572},
  year={2020}
}

@article{plantinga2021perceptual,
  title={Perceptual Loss with Recognition Model for Single-Channel Enhancement and Robust {ASR}},
  author={Plantinga, Peter and Bagchi, Deblin and Fosler-Lussier, Eric},
  journal={arXiv preprint arXiv:2112.06068},
  year={2021}
}

@article{dnsmos_pesq_flaw,
  title={A Scalable Noisy Speech Dataset and Online Subjective Test Framework},
  author={Reddy, Chandan KA and Beyrami, Ebrahim and Pool, Jamie and Cutler, Ross and Srinivasan, Sriram and Gehrke, Johannes},
  journal={Proc. Interspeech},
  pages={1816--1820},
  year={2019}
}

@article{correlates_breathy,
author = {Hillenbrand, James and Cleveland, Ronald and Erickson, Robert},
year = {1994},
month = {09},
pages = {769-78},
title = {Acoustic Correlates of Breathy Vocal Quality},
volume = {37},
journal = {Journal of speech and hearing research},
doi = {10.1044/jshr.3704.769}
}

@article{correlates_breathy_rough,
  title={Some spectral correlates of pathological breathy and rough voice quality for different types of vowel fragments},
  author={Krom, Guus de},
  journal={Journal of Speech, Language, and Hearing Research},
  volume={38},
  number={4},
  pages={794--811},
  year={1995},
  publisher={ASHA}
}

@article{KASUYA1986171,
title = {An acoustic analysis of pathological voice and its application to the evaluation of laryngeal pathology},
journal = {Speech Communication},
doi = {https://doi.org/10.1016/0167-6393(86)90006-3},
url = {https://www.sciencedirect.com/science/article/pii/0167639386900063},
author = {Hideki Kasuya and Shigeki Ogawa and Yoshinobu Kikuchi and Satoshi Ebihara},
  year={1986}
}

\end{document}